# REAL TIME ERROR DETECTION IN METAL ARC WELDING PROCESS USING ARTIFICIAL NEURAL NETWORKS

Prashant Sharma[1], Dr. Shaju K. Albert[2] and S. Rajeswari[3]

[1]Computer Division, Indira Gandhi Centre For Atomic Research, Kalpakkam
prashantcss@igcar.gov.in
[2]Material Technology Division, Indira Gandhi centre for Atomic Research, Kalpakkam
shaju@igcar.gov.in
[3]Computer Division, Indira Gandhi Centre for Atomic Research, Kalpakkam
raj@igcar.gov.in

## ABSTRACT

*Quality assurance in production line demands reliable weld joints. Human made errors is a major cause of faulty production. Promptly Identifying errors in the weld while welding is in progress will decrease the post inspection cost spent on the welding process. Electrical parameters generated during welding, could able to characterize the process efficiently. Parameter values are collected using high speed data acquisition system. Time series analysis tasks such as filtering, pattern recognition etc. are performed over the collected data. Filtering removes the unwanted noisy signal components and pattern recognition task segregate error patterns in the time series based upon similarity, which is performed by Self Organized mapping clustering algorithm. Welder's quality is thus compared by detecting and counting number of error patterns appeared in his parametric time series. Moreover, Self Organized mapping algorithm provides the database in which patterns are segregated into two classes either desirable or undesirable. Database thus generated is used to train the classification algorithms, and thereby automating the real time error detection task. Multi Layer Perceptron and Radial basis function are the two classification algorithms used, and their performance has been compared based on metrics such as specificity, sensitivity, accuracy and time required in training.*

## 1. INTRODUCTION

Manual Arc welding process, though carried out using specifically designed power sources, is a dynamic and stochastic process due to random behavior of the electric arc and the metal transfer that takes place during welding. Fluctuations in voltage and current are so rapid and random that a high speed data acquisition system is required to capture these variations. Data thus, acquired can be analyzed to derive characteristic feature of the welding process. Data collected is huge and contain features corresponding to the performance of welding power source, welding consumables and the welder. As far as power source and welding consumables are concerned, their qualities can be improved by using better technology and better composition respectively. But process is greatly affected by human made errors. Number of errors as well as uniformity of the weld greatly depends upon the experience of welder. Presently existing techniques declare quality of welder by inspecting the weld done by him, through various NDT techniques, which are time consuming and resource intensive. Amit kumar et al. has done a detailed study on utilization of ANN in welding technology[1] and suggest ANN can be used to greatly optimize welding techniques.

T. Polte and D. Rehfeldt [2] suggest better alternative for declaring weld quality. Electrical parameters (current and voltage) obtained from power supply terminal are of random character and varies over wide range. Consequently statistical behavior of these parameters is used to characterize process, which can thereafter be used to detect the type of errors in the process. PDD transformation of voltage time series is fed to the ANN to classify the type of errors. On the other hand, J. Mirapeix and P.B. Garcia-Allende[3] used spectroscopic analysis of plasma spectra produced during welding to monitor quality of resulting weld seams. Plasma spectra captured during welding process goes through PCA technique, thereby reducing spectral dimensions and consequently they are fed to ANN for fault detection task. ANN is also used by Hakan Ates[4] for prediction of welding parameters such as hardness, tensile strength, elongation and impact strength. A. Sanchez Roca obtained a model using ANN for estimating stability of gas metal arc welding [5]. Application of ANNs for prediction of the weld bead geometry using features derived from the infrared thermal video of a welding process has also been done [6].

A novel and less computation intensive method of error detection is presented in this paper. Since the electrical parameters are clear description of the process in real time, so rather than using their PDD transformations we can directly use these waveforms to promptly recognize the process errors. Artificial neural networks (ANN) are employed to analyze the processed data. Self organized mapping (SOM) [9] algorithm clusters the input voltage data based upon waveform similarity. Database thus prepared is used to rank welders as well as it also trains classification algorithms. Multi Layer Perceptron (MLP) and Radial Basis function (RBF) are two type of ANN classifiers used to perform classification task on database obtained from SOM algorithm. Gradient descent algorithm trains MLP model. In RBF, transformation to higher dimension is performed using K means clustering algorithm [8], thereafter learning of separation boundary is done using gradient descent algorithm. Both the ANN classifiers are able to perform classification task fairly accurately. Performance of both has been compared.

## 2. EXPERIMENTAL

Aim of the experiment is to acquire voltage and current values from terminals of power supply. Fig shows block diagram of the setup deployed to perform data acquisition task.

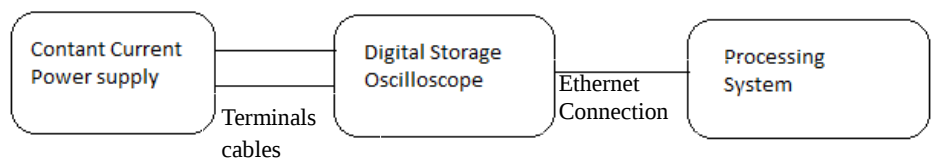

Figure 1. Experimental Setup

Online monitoring of the welding process is done and measured signals are transferred to process computer via Digital Signal Oscilloscope. Hall Effect current sensor measures current though power cord and voltage differential probe measures voltage values across power supply terminals. Voltage and Current signals thus acquired can be considered as time series and they can be subjected to various time series analysis.

Constant current power supply subside the process variations in current signal thus only voltage time series is selected for further processing. Data processing steps such as filtering, down sampling and segmentation was applied to voltage time series and Figure 2 depicts every processing step.

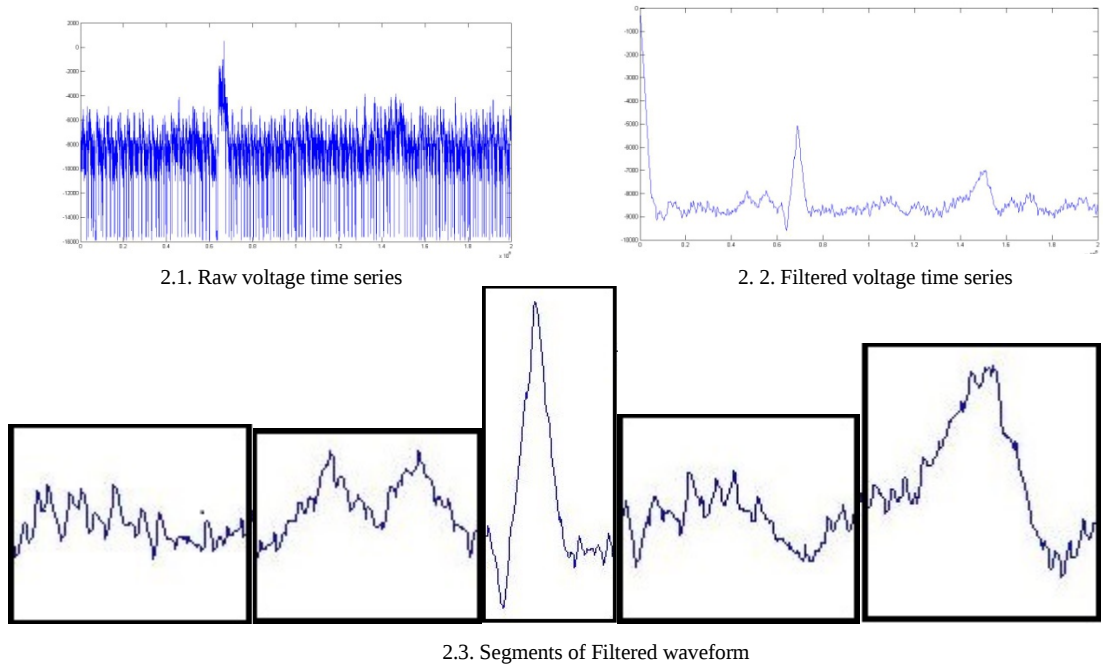

2.1. Raw voltage time series       2. 2. Filtered voltage time series

2.3. Segments of Filtered waveform

Figure 2. figure 2.1 shows raw voltage waveform 2.2 shows filtered waveform and 2.3 shows segments of filtered waveform

In Figure 2.3 third and fifth segments are most unsteady and these patterns correspond to error in weld at that particular time. Detecting these patterns with data mining algorithm will help us to pinpoint error in the weld in real time.

Every welder undergoes 3 trails, thus for 30 welders we are having 90 voltage time series. After applying proper filtering technique, every voltage time series is segmented into 17 segments, each of 100000 data points (thus each welder is having total of 51 segments from his three voltage time series).This procedure give us total of 1530 segments, each of 1,00,000 data points. Thereafter applying downsampling reduces data point in each segment from 100000 to just 50. As signal has only low frequency contents, thus this much downsampling doesn't affect overall shape of segment.

## 3. RESULTS AND DISCUSSIONS

Database of segment is not classified i.e. it is not known that which pattern belong to which class, thus an unsupervised clustering algorithm is required to group together the patterns based on similarity. Therefore, using Self Organized Mapping unsupervised clustering algorithm similar patterns are grouped together to form 9 different groups. Groups that are clustering steady patterns are identified and all the patterns clustered under it are marked as desirable patterns. And patterns under other groups are marked as undesirable. As we already know that each welder has 51 patterns, counting is done for number of undesirable pattern. Welder which has most number of undesirable patterns is ranked the lowest and that is how the ranking is performed.

Table 1. Parameters of SOM algorithm

Table 2. Number of patterns clustered by each clusters

| Parameter | Value |
|---|---|
| Number of training iteration | Until no considerable change in weight of clusters occurs |
| Number of training patterns | 1530 |
| Dimension of each pattern | 50 |
| Initial learning rate | 0.3 |
| Current radius | 5.0 |
| Rate of decreasing in radius | 0.1 |
| Number of clusters | 9 |

| Cluster index | No. of Patterns clustered |
|---|---|
| 1 | 101 |
| 2 | 135 |
| 3 | 232 |
| 4 | 243 |
| 5 | 533 |
| 6 | 81 |
| 7 | 46 |
| 8 | 101 |
| 9 | 58 |

Weights of the cluster 4 and 5 came out to have least standard deviation. Thus all the patterns under it are marked as desirable. And patterns under other clusters are marked as undesirable and these are error patters. Now, the pattern database is classified and is suitable for supervised learning of classification algorithm. This step will help in automating the task of error detection.

Database is now modified by marking 1 against the pattern which is belonging to desirable class and marking 0 against the patterns that belongs to undesirable class. This modified database is used for supervised learning of Multilayer Perceptron and radial basis function type neural networks. 70% of database is used for learning and 30% is used for testing. Comparative study of performance of both networks on same database is done.

Table 3. Parameter values for MLP

| Parameter | Values |
|---|---|
| Number of training iteration | 10000 |
| Number of hidden layer | 1 and 2 |
| Initial learning rate | 0.3 |
| Rate of decreasing learning rate | 0.001 |

Table 4. Parameter values for RBF

| Parameter | Values |
|---|---|
| Number of training iteration | 10000 |
| Number of hidden layer | 1 |
| Initial learning rate | 0.3 |

| Regularization parameter | 0.3 |

### 3.1. SIMULATION RESULTS

Training of classification algorithms is done using 1021 entries and last 509 entries are used for testing purpose. Selection of number of hidden nodes in single layer and double layer MLP was done based on study done by Wanas [7].

Table 6. Topology of MLP and RBF networks

| | Topology | True classification | misclassification | %correct evaluation |
|---|---|---|---|---|
| MLP | 50-25-25-2 | 477 | 32 | 93.71 |
| | 50-35-2 | 458 | 51 | 89.99 |
| RBF | 50-95-2 | 455 | 54 | 89.31 |
| | 50-80-2 | 452 | 57 | 88.80 |

Table 7. Comparison between RBF and MLP

| | MULTILAYER PERCEPTRON | | RADIAL BASIS NETWORK | |
|---|---|---|---|---|
| No of nodes | 35 nodes | 25 nodes Layer1 25 nodes Layer2 | 80 nodes | 95 nodes |
| Sensitivity | 0.8755 | 0.9572 | 0.8055 | 0.8288 |
| Specificity | 0.9246 | 0.9365 | 0.9762 | 0.9603 |
| Accuracy% | 89.9804 | 94.6955 | 88.8016 | 89.391 |
| Time required | 9mins 23secs | 16 mins 10 secs | 1min 52secs | 2 mins 9 sec |

Table 6 suggest that 2 layer MLP network gave least % test error among the three types of networks. % test error shown by RBF network was indeed highest, but time required for training the RBF network was 5 times less as compared to that required by MLP network as shown in Table 7. It is also seen that, specificity, which is the fraction of true negative classifications, was highest for RBF network and sensitivity, which is the fraction of true positive classification, was highest for two layer network. Comparison results between MLP and RBF are complying with the study done by Santos et al. [10].

Choosing any of the two trained neural network (based upon the accuracy needed) we could able to detect number of error patterns in the incoming voltage time series in real time, thereby detecting the errors in weld in real time.

### 4. CONCLUSIONS:

Technique is developed to detect number of error patterns in a time series using Artificial Neural Networks. Self Organized mapping algorithm could successfully segregate the steady patterns and patterns that are in error. Reference database is generated using clustering process done by SOM. Both MLP and RBF network can easily be trained with reference database and thereby be used to classify an unknown pattern. Two layered MLP gave better error performance than RBF but training time required by MLP was almost 5 times as that required by RBF.